\documentclass[a4paper,twoside]{article}

\usepackage{epsfig}
\usepackage{subfigure}
\usepackage{calc}
\usepackage{amssymb}
\usepackage{amstext}
\usepackage{amsmath}
\usepackage{amsthm}
\usepackage{multicol}
\usepackage{pslatex}
\usepackage{apalike}
\usepackage{SCITEPRESS}     

\usepackage{xcolor}

\graphicspath{{figures/}}

\begin{document}

\title{DeepBall: Deep Neural-Network Ball Detector}

\author{\authorname{Jacek Komorowski \sup{1, 2}, Grzegorz Kurzejamski \sup{1, 2} and Grzegorz Sarwas \sup{1, 2}}
\affiliation{\sup{1}Warsaw University of Technology, Warsaw, Poland}
\affiliation{\sup{2}Sport Algorithmics and Gaming Sp. z o.o., Warsaw, Poland}
\email{jacek.komorowski@pw.edu.pl, g.kurzejamski@sagsport.com, sarwasg@ee.pw.edu.pl}
}


\abstract{The paper describes a deep network based object detector specialized for ball detection in long shot videos. Due to its fully convolutional design, the method operates on images of any size and produces \emph{ball confidence map} encoding the position of detected ball. The network uses hypercolumn concept, where feature maps from different hierarchy levels of the deep convolutional network are combined and jointly fed to the convolutional classification layer. This allows boosting the detection accuracy as larger visual context around the object of interest is taken into account. The method achieves state-of-the-art results when tested on publicly available ISSIA-CNR Soccer Dataset.}

\onecolumn \maketitle \normalsize \vfill

\section{\uppercase{Introduction}}
\label{sec:introduction}

\noindent 

An ability to accurately detect and track the ball in a video sequence is a core capability of any system aiming to automate analysis of the football matches or players' progress. Our method aims to solve the problem of fast and  accurate ball detection. It is developed as a part of the computer system for football clubs and academies to track and analyze player performance during both training session and regular games. The system is intended to help professional football analysts to evaluate the players' performance, by allowing automatic indexing and retrieval of interesting events. 

\begin{figure*}
  \centering
  \includegraphics[width=1.0\textwidth]{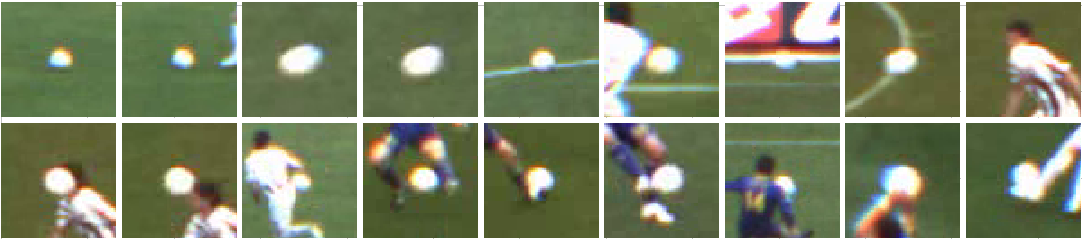}
  \caption{Exemplary patches illustrating high variance in ball appearance and difficulty of the ball detection task.} 
  \label{jk:fig:ball_images}
\end{figure*}

Detecting the ball from long-shot video footage of a football game is not trivial to automate. The object of interest (the ball) has very small size compared to other objects visible in the observed scene. 
Due to the perspective projection, its size varies depending on the position on the play field. 
The shape is not always circular. When a ball is kicked and moves at high velocity, its image becomes blurry and elliptical.
Perceived colour of the ball changes due to shadows and lighting variation.
The colour is usually similar to the colour of white lines on the pitch and sometimes to players' jerseys.
Other objects with similar appearance to the ball can be visible, such as small regions near the pitch lines and regions of players' bodies such as a head. 
Situations when the ball is in player's possession or partially occluded are especially difficult.
Figure \ref{jk:fig:ball_images} shows exemplary image patches illustrating high variance in the ball appearance and difficulty of the ball detection task. 

Traditional ball detection methods, e.g. based on variants of circular Hough transform, deal well with situations where ball is visible as a single object, separated from the player body. They have problems to detect the ball when it's possessed or partially occluded by a player. But for players performance analysis purposes, the most informative are frames showing players in close contact with the ball.
In this paper we present a ball detection method expanding upon the state-of-the-art deep convolutional object detection network. 
The method operates on a single video frame and is intended as the first stage in the ball tracking pipeline. 
Our method does not have limitations associated with earlier methods based on a circular Hough transform.  
It can deal with situations where the perceived ball shape is not circular due to the motion blur. It detects the ball when it's in a close contact with or partially occlude by a player's body. It can detect multiple balls, located relatively close to each other, in the same image.
Another benefit of the proposed method is its flexibility. Due to the fully convolutional design it can operate on images of any size and produces the ball confidence map of a size proportional to the input image.
The detection network is designed with performance in mind.
Evaluation performed in Section \ref{jk:ev_results} proves that our method can efficiently process high definition video input in a real time.

\section{\uppercase{Related Work}}
\noindent
The first step in the traditional ball detection methods, is usually the process of background subtraction. It prevents ball detection algorithms from producing false detections on the static part of the image such as stadium advertisement. 
The most commonly used background subtraction approaches are based on chromatic 
features~\cite{Gong95,Ali12,Kia16} or motion detection~\cite{DOr02,DOr04,Leo08,Mazz12}.
Segmentation methods based on chromatic features use domain knowledge about the visible scene: football pitch is mostly green and the ball mostly white.  The colour of the pitch is usually modelled using a Gaussian Mixture Model and hardcoded in the system or learned. When the video comes from the static camera, motion-based segmentation is often used. 
For computational performance reasons, a simple approach is usually applied based on an absolute difference between consecutive frames or the difference between the current frame and the mean or median image obtained from a few previously processed frames \cite{High16}.

After the background segmentation, heuristic criteria based on chromatic or morphological features are applied on the resulting blobs to locate the ball. These criteria include blob size, colour and shape (circularity, eccentricity)~\cite{Gong95}. Variants of Circle Hough Transform~\cite{Yuen90}, modified to detect spherical rather than circular objects, may be used to verify if a blob contains the ball ~\cite{DOr02,DOr04,Leo08,Popp10,Halb15}.
A two-stage approach may be employed to achieve real-time performance and high detection accuracy~\cite{DOr02,Leo08,Mazz12}. In this scenario the regions that probably contain the ball are found (\emph{ball candidates extraction}). Then, the candidates are validated (\emph{ball candidate validation}). 

In~\cite{Ali12} straight lines are detected using kernel-based Hough transform and removed from the foreground image to overcome problem of ball interfusing with white lines on the pitch. Very similar method is proposed in \cite{Rao15}. 
~\cite{Gong95,Pall08,Halb15} use multiple successive frames to improve the detection accuracy.
In \cite{Gong95}, detection is confirmed by searching a neighbourhood area of each ball candidate in the successive frame. If the white area with similar size and circularity is found in the next frame, the ball candidate is validated. 
In \cite{Pall08} authors extract ball candidate positions using morphological features (shape and size of the ball). Then, a directed weighted graph is constructed from ball candidates in successive frames. The vertices of the graph correspond to candidate ball positions and edges link candidates found in consecutive frames. The longest path in the graph is computed to give the ball trajectory.

Ball detection methods using morphological features to analyze shape of blobs produced by background segmentation, fail if a ball is touching a player. See bottom row of Fig.~\ref{jk:fig:ball_images} for exemplary images where these methods are likely to fail. 
\cite{Halb15} addresses this limitation by using two-stage approach. First, the ball is detected in not occluded situations, where it appears as a single object. This is done by applying background subtraction to filter out temporally static part of the image. Then, foreground blobs are filtered by size and shape to produce ball candidates. Ball candidates are verified by examining a few successive frames and detecting robust partial ball trajectories (tracklets). When the first stage detector is not able to locate the ball, the second stage detector specialized for partially occluded situations is used.
Ball candidates are found using a Hough circle detector. Foreground object contours are extracted and their Freeman chain code is examined. If a ball candidate corresponds to a 'bump' in the foreground object silhouette it is retained as a true match.


In recent years a significant progress was made in the area of neural-network based object detection. 
Deep neural-network based YOLO detector~\cite{Redm16} achieves 63.4 mean Average Precision (mAP) on PASCAL VOC 2007 dataset, whereas traditional Deformable  Parts  Models (DPM) detector~\cite{Felz10} scores only 30.4. Current state-of-the-art object detectors can be categorized as one-stage or two-stage. In two-stage detector, such as: Fast R-CNN~\cite{Girs15} or Faster R-CNN~\cite{Ren15},
the first stage generates a sparse set of candidate object locations (region proposals). The second stage uses deep convolutional neural network to classify each candidate location as one of the foreground classes or as a background. One-stage detectors, RetinaNet~\cite{Lin17}, SSD~\cite{Liu16} or YOLO~\cite{Redm16}, do not include a separate region-proposal generation step. A single detector based on deep convolutional neural network is applied instead. 
 
\cite{Spec17} uses convolutional neural networks (CNN) to localize the ball under varying environmental conditions. The first part of the network consists of multiple convolution and max-pooling layers which are trained on the standard object classification task. The output of this part is processed by fully connected layers regressing the ball location as probability distribution along x- and y-axis. The network is trained on a large dataset of images with annotated ground truth ball position. The network is reported to have 87\% detection accuracy on the custom made dataset. The limitation of this method is that it fails if more than one ball, or object very similar to the ball, is present in the image. Our method does not have this limitation.

\cite{Reno18} presents a 
deep neural network classifier, consisting of convolutional feature extraction layers followed by fully connected classification layer. It is trained to classify small, rectangular image patches as ball or no-ball.
The classifier is used in a sliding window manner to generate a probability map of the ball occurrence. 
The method has two drawbacks. 
First, the set of negative training examples (patches without the ball) must be carefully chosen to include sufficiently hard examples. 
Also the rectangular patch size must be manually selected to take into account all the possible ways the ball appears on the scene: big or small due to the perspective, sharp or blurred due to its speed. 
The method is also not optimal from the performance perspective.
Each rectangular image patch is separately processed by the neural network using a sliding-window approach. Then, individual results are combined to produce a final ball probability map.
Our method, in contrast, requires only a single pass of an entire image through the fully convolutional detection network. 




\section{\uppercase{Deep network-based ball detection method}}

The method presented in this paper, called \emph{DeepBall}, is inspired by recent advances in a single-pass deep neural network based object detection methods, such as SSD~\cite{Liu16} or YOLO~\cite{Redm16}.
A typical architecture of a neural network-based one stage object detector is modified, to make it more appropriate for the ball detection task. Modifications aim at increasing accuracy of locating small objects and reducing the processing time. The network is designed to take larger visual context into the consideration to correctly classify fragments of the scene containing objects similar to the ball. This is achieved by using hypercolumn concept introduced in \cite{Hari15}. 
In order to increase the performance, we removed unnecessary components typical for single stage neural network object detector.
Multiple anchor boxes, with different size and aspect ratios, are not needed as we detect objects from a single class (the ball) with a limited shape and size variance.
Localization module, predicting the centre and size of object bounding boxes relative to a grid cell is unnecessary, as proposed method produces a dense confidence map predicting the ball location on a pixel level. 

\begin{figure}
  \includegraphics[width=3.3cm, trim={5cm 2cm 3cm 0},clip]{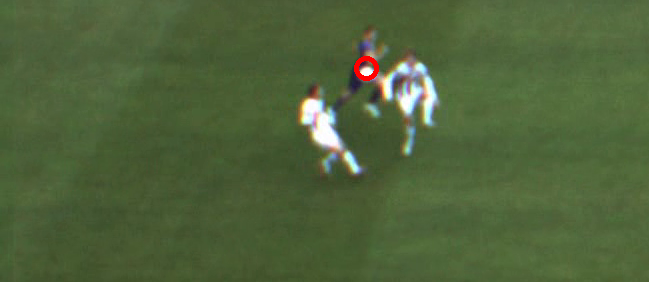}
  \includegraphics[width=3.3cm, trim={5cm 2cm 3cm 0},clip]{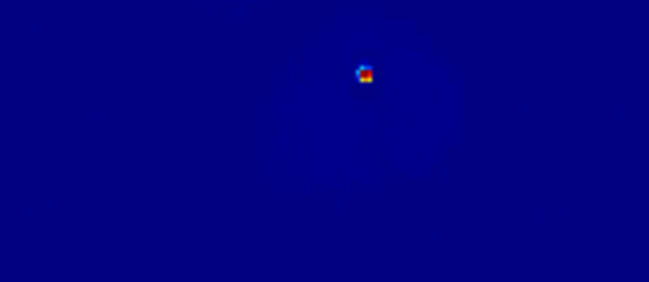}
  \includegraphics[height=1.9cm]{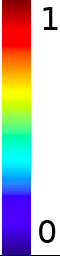}
  \caption{Part of the exemplary input frame from the test sequence with highlighted ball position (left) and corresponding \emph{ball confidence map} (right)}
  \label{jk:fig:input_output}
\end{figure}

The method takes a video frame of any resolution as an input and produces scaled down \emph{ball confidence map} encoding probability of ball presence at each location.
The size of the output \emph{ball confidence map} is  $h_f \times w_f$, where $h_f$ and $w_f$ equal to the original image height and width divided by the scaling  factor $k$ ($k=4$ in our case).
Position in the \emph{ball confidence map} with coordinates $(x_f, y_f)$ corresponds to the position 
$(\lfloor k(x_f-0.5) \rfloor, \lfloor k(y_f-0.5) \rfloor$ in the input image.
See Fig.~\ref{jk:fig:input_output} for an exemplary input image and corresponding \emph{ball confidence map} computed by the trained network.
The actual ball position is retrieved from the \emph{confidence map} using the following approach. 
First, the location with the highest confidence is found in the \emph{ball confidence map}.
If the confidence is lower than a threshold $\theta$, no balls are detected.
Otherwise, the location with the highest confidence is returned.
In 'training game mode', where more than one ball can be present in the image, more balls are detected. 
This is done by zeroing-out confidence map values at the previously found maximum and its close neighbourhood  (non-max suppression) and searching for the second global maximum.
The process is repeated until no new maximum with confidence above the threshold $\theta$ can be found.
Pixel coordinates of the ball $(x_p, y_p)$ in the input frame are calculated using the following formula:
$(x_p, y_p) = (\lfloor k(x_f-0.5) \rfloor, \lfloor k(y_f-0.5) \rfloor$, 
where $(x_f, y_f)$ are coordinates in the \emph{ball confidence map} with the maximum confidence and $k=4$ is a scaling factor.
The threshold $\theta$ is set experimentally, as the value maximizing detection accuracy on the validation set.

\paragraph{Network architecture}

\begin{figure*}
 \centering
  \includegraphics[clip, trim=2cm 5cm 1.5cm 2.5cm, width=1.0\textwidth]{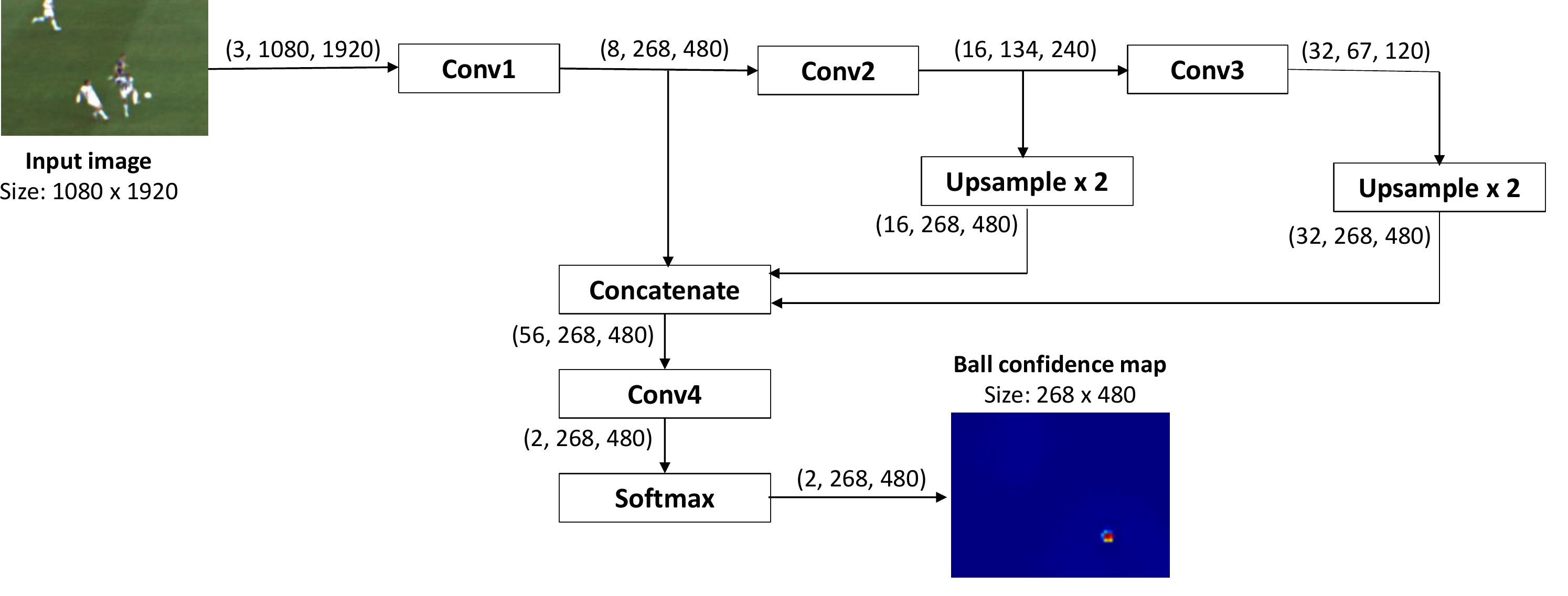}
    \caption{High-level architecture of \emph{DeepBall} network.
    The input image is processed by three convolutional blocks (Conv1, Conv2 and Conv3) producing convolutional feature maps with decreasing spatial resolution and increasing number of channels.
    Feature maps are upsampled to the same spatial resolution and concatenated along channels dimension. Concatenated feature map is fed to the final fully convolutional classification block (Conv4) followed by Softmax. The output is two channel \emph{ball confidence map}.
    }
  \label{jk:fig:network-diagram}
\end{figure*}

\begin{table}[ht]
\caption{Detailes of \emph{DeepBall} network architecture. Output size is specified in the format: (number of channels, height, width).
Each convolutional layer is followed by BatchNorm layer and ReLU non-linearity (not show for brevity).
All convolutions use same padding and stride one (except for the first one).
}
\begin{center}
\begin{tabular}{l@{\quad}l@{\quad}l@{\quad}l}
\hline
Block & Layers & Output size \\
 \hline
Conv1  &  Conv: 8 7x7 filters &  \\
       &                  stride 2 & \\
       & Conv: 8 3x3 filters &  \\
       & Max pool: 2x2 filter & (8, 268, 480)\\
Conv2  &  Conv: 16 3x3 filters    &   \\
       & Conv: 16 3x3 filters &  \\
       & Max pool: 2x2 filter  & (16, 134, 240)\\
Conv3  &  Conv: 32 3x3 filters   &   \\
       & Conv: 32 3x3 filters & \\
       & Max pool: 2x2 filter  & (32, 67, 120)\\
Conv4   &  Conv: 56 3x3 filters    &  \\
        &  Conv: 2 3x3 Filters    &  (2, 268, 480) \\
Softmax &  Softmax & (2, 268, 480) \\
[2pt]
\hline
\end{tabular}
\end{center}
\label{jk:table2}
\end{table}

The diagram depicted in Fig.~\ref{jk:fig:network-diagram} shows components of our ball detection network and size of outputs of each block.
Note that output size depends on the size of the input image, as the network is fully convolutional and can operate on the image of any size.
The input image is processed by three convolutional blocks (Conv1, Conv2 and Conv3) producing convolutional feature maps with decreasing spatial resolution and increasing number of channels.
In contrast to a typical convolutional network design, the output from each convolutional block is concatenated and jointly fed into the final classification layer.
Feature maps produced by convolutional blocks Conv2 and Conv3 are first upsampled to the same spatial resolution as a feature map produced by the first convolutional block (Conv1).
Then, the feature map produced by the first convolutional block (Conv1) and upsampled feature maps from second and third convolutional blocks (Conv2 and Conv3) are concatenated along the dimension corresponding to the number of channels to form a hypercolumn.
Concatenated feature map is fed to the final fully convolutional classification block (Conv4).
The classification block consists of two convolutional layers followed by the softmax layer. It outputs two channel \emph{ball confidence map}. One channel is interpreted as the probability of the location being a background and the other as probability of the ball. For the ball detection task, one output channel, interpreted as the ball probability, would be sufficient. But the proposed design is extensible and can be easily adapted to  accommodate detection of additional object categories, such as players.
Detailed architecture of each block is given in Table~\ref{jk:table2}.

Concatenation of multiple convolutional feature maps from different level of the network, allows using both low-level features from the first convolutional layers and high-level features computed by higher convolutional layers.
Information from first convolutional layers is necessary for a precise spatial location of the object of interest.
Further convolutional layers operate on feature maps with lower spatial resolution, thus they cannot provide exact spatial location. 
But they have bigger receptive fields and their output can provide additional context to improve classification accuracy.
This design is inspired by the hypercolumn concept~\cite{Hari15}, where outputs from intermediary convolutional layers are upsampled and concatenated in order to allow find-grained object localization.


The network architecture described above was chosen experimentally by evaluating a number of alternative designs. See Section~\ref{jk:section-experimental-results} for information on examined variants and their performance.

\paragraph{Loss function} is a modified version of the loss used in SSD~\cite{Liu16} detector. Proposed network does not regress position and size of the object's bounding box. The ball position is determined by the maxima of the confidence map computed by the network. Hence only the classification component of the original SSD loss function is used. The loss $\mathcal{L}$ optimized during the training is cross-entropy loss over ball and background class confidences:
\begin{equation}
\begin{aligned}
\mathcal{L} \left( c \right) = \frac{1}{N} 
\left(-\sum_{\left(i,j\right)\in Pos}\log\left(c_{ij}^{ball}\right) \right. \\
\left.-\sum_{\left(i,j\right)\in Neg}\log\left(c_{ij}^{bg}\right) \right),
\end{aligned}
\end{equation}
where 
$c_{ij}^{bg}$ is the value of the channel of the ball confidence map corresponding to the background probability at the spatial location $(i, j)$ and $c_{ij}^{ball}$ is the is the value of the channel of the ball confidence map corresponding to the ball probability at the spatial location $(i, j)$. $Pos$ is a set of positive examples, that is the set of spatial locations on the ball confidence map corresponding to the ground truth ball location. $Neg$ is a set of negative examples, that is the set of spatial locations on the ball confidence map corresponding to the ground truth background.

Set of positive examples $Pos$ is constructed as follows. If $(x,y)$ is a true ball position for the image $I$, then the corresponding confidence map location $(i,j) = ( \lfloor x/4, y/4 \rfloor )$ and all its nearest neighbours are added to $Pos$.

Negative examples (locations without the ball) correspond to locations on the confidence map, where the ball, according to the ground truth data, is not present.
The number of negative examples is orders of magnitude higher than a number of positive examples (locations with the ball) and this would create highly imbalanced training set.
To mitigate this, we employ hard negative mining strategy as in~\cite{Liu16}. We chose a limited number of negative examples with the highest confidence loss, so the ratio of negative to positive examples is at most 3:1.

\paragraph{Training dataset}
\emph{DeepBall} network is trained using the publicly available ISSIA-CNR Soccer Dataset \cite{DOr09}. The dataset contains six synchronized, long shot views of the football pitch acquired by six Full-HD DALSA 25-2M30 cameras. Three cameras are designated for each side of the playing-field, recording at 25 fps. Videos are acquired during matches of the Italian 'serie A'. There're 20,000 manually annotated frames in the dataset, out of which 7,000 contain the ball and 13,000 doesn't or the ball is occluded by players. The ball radius varies from 8 to 16 pixels. Sequences 1, 2, 3 and 4, covering one penalty area and the centre of the football pitch, are used for training. Sequences 5 and 6, covering the side of football pitch not visible on the training sequences, are left aside for the evaluation purposes.

\begin{figure}
  \centering
  \includegraphics[width=0.45\textwidth]{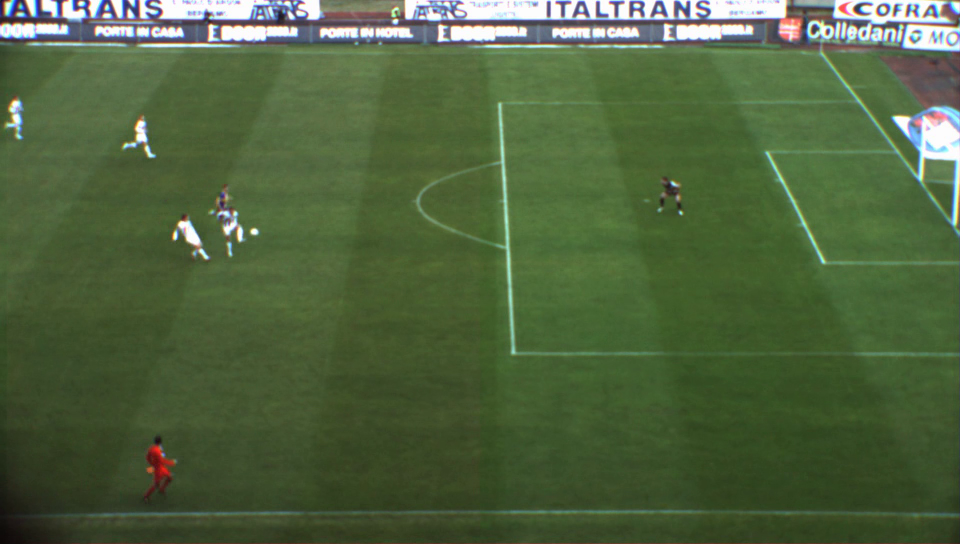}
    \caption{Exemplary frame from the training dataset.}
  \label{jk:fig:training_sequences}
\end{figure}

Fig. \ref{jk:fig:training_sequences} shows exemplary frames from the sequence 1 and 3. As the training dataset is relatively small, we use data augmentation to increase the variety of training examples and decrease the risk of overfitting. The following transformations are randomly applied to the training images: random color jitter (random change in brightness, contrast, saturation or hue), horizontal flip, random cropping and random scaling (with scale factor between 0.5 and 1.1). The ground truth (ball position) is modified accordingly to align with the transformed image.

The network is trained using a standard gradient descent approach with Adam~\cite{King14} optimizer. The initial learning rate is set to $0.001$ and decreased by 10 after 50 epochs. The training runs for 75 epochs in total. Batch size is set to 16.

\section{Experimental results}
\label{jk:section-experimental-results}

\paragraph{Evaluation dataset}
Evaluation is performed on two datasets. The first contains of sequence 5 and 6 from the ISSIA-CNR Soccer Dataset. This sequence covers the part of the football pitch not seen on the training sequences (sequence 1, 2, 3 and 4). ISSIA-CNR dataset is quite demanding because the video has a moderate quality and there's noticeable blur. One of the team wears white jerseys which makes difficult to  distinguish the ball when it's close to the player.




\paragraph{Evaluation metrics}
\label{sec:metrics}
We evaluate Average Precision (AP), a standard metric used in assessment of object detection methods. We follow Average Precision definition from Pascal 2007 VOC Challenge \cite{Ever10}.
The precision/recall curve is computed from a method’s ranked output. Recall is defined as a proportion of all positive examples ranked above a given threshold to all positive examples in the ground truth. Precision is a proportion of all positive examples above that threshold to all examples above that threshold. The AP summarizes the shape of the precision/recall curve, and is defined as the mean precision at a set of eleven equally spaced recall levels:

\begin{equation}
\mathrm{AP} = \frac{1}{11} \sum_{r \in \left\{ 0, 0.1, \ldots 1 \right\}} p(r) \; ,\end{equation} 
where $p(r)$ is a precision at recall level $r$.

The ball detection method usually operates under the additional constraint, that no more than one object of interest (the ball) is present in the image. Under this constraint, for each image the detector returns the highest response from the ball confidence map greater than the threshold $\theta$ as the ball position. If no location in the ball confidence map is greater than $\theta$, no ball is detected. In this scenario, an image with the ball is classified correctly, if the ball is detected at the correct location. The image without the ball is classified correctly, if ball is not detected. \emph{Ball detection accuracy} is defined as the proportion of correctly classified images to all processed images. $\theta$ is chosen experimentally, as the value maximizing the accuracy on the validation set.

\paragraph{Evaluation results}
\label{jk:ev_results}
Evaluation results are summarized in Table~\ref{jk:table1}. The results contain Average Precision and Accuracy of evaluated methods, as defined in the previous section. 
The table also lists a number of trainable parameters in each evaluated model and frame rate, expressed in frames per second, achievable when detecting the ball in a Full HD (1920x1080 resolution) video.
Frame rates given in the table take into account the time needed to feed a frame through the detection network and infer the ball position from the resultant feature map. 
They do not include the time needed to load the frame from an input file, convert it to the tensor and load into the GPU. All methods are implemented in PyTorch~\cite{Pasz17} and run on nVidia Titan X GPU platform.

Our method yields the best results on the test set (Sequences 5 and 6 from ISSIA-CNR Soccer Dataset). It achieves 0.877 Average Precision and 0.951 ball detection accuracy. 
For comparison we evaluate two recent ball detection methods: \cite{Spec17} and \cite{Reno18} using the same training and test sets and the same data augmentation approach as in our method. 

\cite{Spec17} uses the neural network with three convolutional layers followed by two two-layer fully connected heads estimating the ball x and y coordinates. 
For evaluation we implemented the best performing model proposed in the paper: Model 1 soft-sign. The model performs poorly on the test dataset, achieving only 0.220 Average Precision. 
This can be attributed to the fact, that the original model is intended to detect the ball in videos from RoboCup Soccer matches taken from closer distance. The ball image is larger and there are no visible distractors such as advertisement stands around the pitch.
The method regresses only one ball position on the input image. If there are multiple objects with ball-like appearance, it likely gets confused and fails to produce the meaningful result.
Our method computes a dense confidence map indicating probable ball positions.
It's more robust against presence of objects with similar appearance to the ball.

\cite{Reno18} uses the network consisting of four convolutional layers followed by a fully connected classification layer. This method scores 0.834 Average Precision and 0.917 accuracy. In contrast to the original method, we enhanced the training set construction process. Negative examples (no ball patches) do not need to be manually selected. They are mined online during the network training, as regions of the image not containing the ball but incorrectly classified with the highest confidence (hard negative mining). 
Even with this improvement, the method yields worse Average Precision and detection accuracy than our method.

It must be noted that, while our method outperforms two other neural network based ball detection methods in terms of average precision and detection accuracy, it has significantly lower number of trainable parameters and much higher video processing rate (FPS).

\begin{table*}[t]

\caption{Ball detection method evaluation results.
}
\begin{center}
\begin{tabular}{l@{\quad}c@{\quad}c@{\quad}r@{\quad}r@{\quad}l}
\hline
Method & \begin{tabular}{@{}c@{}}Average\\Precision\end{tabular} & Accuracy & \begin{tabular}{@{}c@{}}No. of trainable\\parameters\end{tabular} & FPS
\\
\hline
DeepBall  &  \textbf{0.877}  &  \textbf{0.951}  & 48 658 & 190 \\
DeepBall (no data augmentation) &  0.792 &  0.899 & 48 658 & 190 \\
DeepBall (no hypercolumns/context)&  0.833  &  0.911 & 29 146 & 270\\
\hline
\cite{Spec17} &  0.220  & 0.220  & 332 365 744 & 22 \\
\cite{Reno18} &  0.834  &  0.917 & 313 922 & 32 \\
[2pt]
\hline
\end{tabular}
\end{center}
\label{jk:table1}
\end{table*}

\begin{figure}
  \centering
  \includegraphics[width=0.15\textwidth]{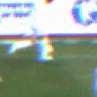} 
  \includegraphics[width=0.15\textwidth]{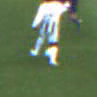}
  \includegraphics[width=0.15\textwidth]{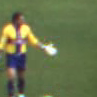} \\
  \includegraphics[width=0.15\textwidth]{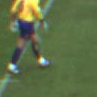}
  \includegraphics[width=0.15\textwidth]{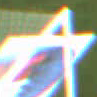}
  \includegraphics[width=0.15\textwidth]{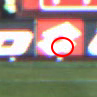}
  \caption{Visualization of incorrect detection results. Top row show image patches where the ball is not detected (false negatives). The bottom row shows patches with incorrectly detected ball (false positives).}
  \label{jk:fig:misclassifications}
\end{figure}

Due to the relatively small size of the training set, data augmentation proved to be the key allowing generalization of the trained network and good performance on the testing set. Without data augmentation Average Precision drops down from 0.877 to 0.792.

Implementing hypercolumn concept by combining convolutional feature maps from different levels of the hierarchy have a positive impact on the method performance. Using a network with a simpler architecture, which bases classification on the output from the last convolutional layer, without combining multiple feature maps, produces worse results. Such architecture scored only 0.833 Average Precision. 

Fig.~\ref{jk:fig:misclassifications} show examples of incorrect detections. Two top rows show image patches where our method fails to detect the ball (false negatives). It can be noticed, that misclassification is caused by severe occlusion, where only small part of the ball is visible, or due to blending of the ball image with white parts of the player wear or white background objects outside the play field, such as stadium advertisement. The bottom row shows examples of patches where a ball is incorrectly detected (false positives). The detector is sometimes confused by players' white socks or by the background clutter outside the play field.


\section{\uppercase{Conclusions}}
\noindent

The article describes an efficient and effective deep neural network based ball detection method. 
The proposed network has a fully convolutional architecture processing entire image at once, in a single pass through the network. This is much more computationally effective than a sliding window approach proposed in ~\cite{Reno18}. Additionally, the network can operate on images of any size that can differ from size of images used during the training.
It outputs scaled down ball confidence map, indicating estimated ball location. The method performs very well on a challenging ISSIA-CNR Soccer Dataset \cite{DOr09} resulting in 0.877 Average Precision and 0.951 accuracy. 
It outperforms two other, recently proposed, neural network-based ball detections methods: \cite{Spec17} and \cite{Reno18}, while having lower number of trainable parameters and significantly higher frame rate.

In the future we plan to use temporal information to improve the system accuracy. Combining convolutional feature maps from few subsequent frames gives additional information that may help to discriminate static, ball-like objects (e.g. parts of stadium advertisement or spare balls located outside the play field) from the moving ball.


\section*{Acknowledgements}
This work was co-financed by the European Union within the European Regional Development Fund.

\bibliographystyle{apalike}
{\small
\bibliography{db-bib}}

\vfill
\end{document}